\documentclass{article}

\usepackage{arxiv}

\usepackage{cite}
\usepackage{amsmath,amssymb,amsfonts}
\usepackage{graphicx}
\usepackage{textcomp}
\usepackage{xcolor}
\usepackage{bm}
\usepackage{comment}
\usepackage{subcaption}
\usepackage{multirow}
\usepackage{booktabs}
\graphicspath{{figures/}}
\usepackage{wrapfig} 
\usepackage{diagbox} 
\usepackage{hyperref,url}

\usepackage{pifont}

\title{Ranking Feature-Block Importance in Artificial Multiblock Neural Networks}

\author{
 \href{https://orcid.org/0000-0002-6919-3483}{\includegraphics[scale=0.06]{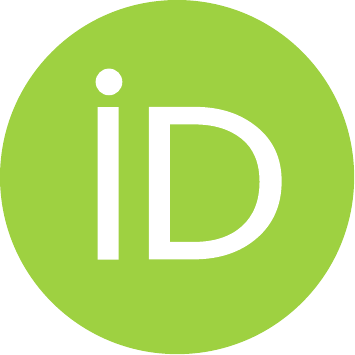}\hspace{1mm}Anna Jenul},
 \href{https://orcid.org/0000-0003-1327-4855}{\includegraphics[scale=0.06]{orcid.pdf}\hspace{1mm}Stefan Schrunner},
 \href{https://orcid.org/0000-0001-5210-132X}{\includegraphics[scale=0.06]{orcid.pdf}\hspace{1mm}Bao Ngoc Huynh},
 \href{https://orcid.org/0000-0001-7455-3108}{\includegraphics[scale=0.06]{orcid.pdf}\hspace{1mm}Runar Helin},\\
 \href{https://orcid.org/0000-0001-7944-0719}{\includegraphics[scale=0.06]{orcid.pdf}\hspace{1mm}\textbf{Cecilia M. Futsæther}},
 \href{https://orcid.org/0000-0001-6468-9423}{\includegraphics[scale=0.06]{orcid.pdf}\hspace{1mm}\textbf{Kristian H. Liland}}, and
 \href{https://orcid.org/0000-0003-1595-9962}{\includegraphics[scale=0.06]{orcid.pdf}\hspace{1mm}\textbf{Oliver Tomic}} \\
  Faculty of Science and Technology\\
  Norwegian University of Life Sciences\\
  \texttt{\{anna.jenul,stefan.schrunner,ngoc.huynh.bao,runar.helin,}\\
  \texttt{cecilia.futsaether,kristian.liland,oliver.tomic\}@nmbu.no} }

\renewcommand{\shorttitle}{Ranking Feature-Block Importance in Multiblock ANNs}

\begin{document}
\maketitle

\begin{abstract}
In artificial neural networks, understanding the contributions of input features on the prediction fosters model explainability and delivers relevant information about the dataset.
While typical setups for feature importance ranking assess input features individually, in this study, we go one step further and rank the importance of groups of features, denoted as feature-blocks.
A feature-block can contain features of a specific type or features derived from a particular source, which are presented to the neural network in separate input branches (multiblock ANNs).
This work presents three methods pursuing distinct strategies to rank features in multiblock ANNs by their importance: (1) a composite strategy building on individual feature importance rankings, (2) a knock-in, and (3) a knock-out strategy.
While the composite strategy builds on state-of-the-art feature importance rankings, knock-in and knock-out strategies evaluate the block as a whole via a mutual information criterion.
Our experiments consist of a simulation study validating all three approaches, followed by a case study on two distinct real-world datasets to compare the strategies.
We conclude that each strategy has its merits for specific application scenarios.
\end{abstract}


\section{Introduction}
In machine learning, datasets with an intrinsic block-wise input structure are common; blocks may represent distinct data sources or features of different types and are frequently present in datasets from industry \cite{dagnely2018}, biology \cite{alnemer2011}, or healthcare \cite{cao2014tensor}. For example, in healthcare, heterogeneous data blocks like patient histology, genetics, clinical data, and image data are combined in outcome prediction models. However, good prediction models do not necessarily depend equally on each block. Instead, some blocks may be redundant or non-informative. Identifying the key data sources in multi-source treatment outcome models promises to deliver new insights into the behavior of black-box models like ANNs. In particular, potential benefits include improving model explainability, reducing costly data acquisitions that do not contribute to the model prediction, and allowing domain experts to explore latent relations in the data. Thus, there is a need to measure the importances of feature-blocks, denoted as feature-block importance ranking (BIR).

\begin{wrapfigure}{r}{0.4\textwidth}
    \centering
    \includegraphics[width=0.45\textwidth, trim=270 0 180 30, clip]{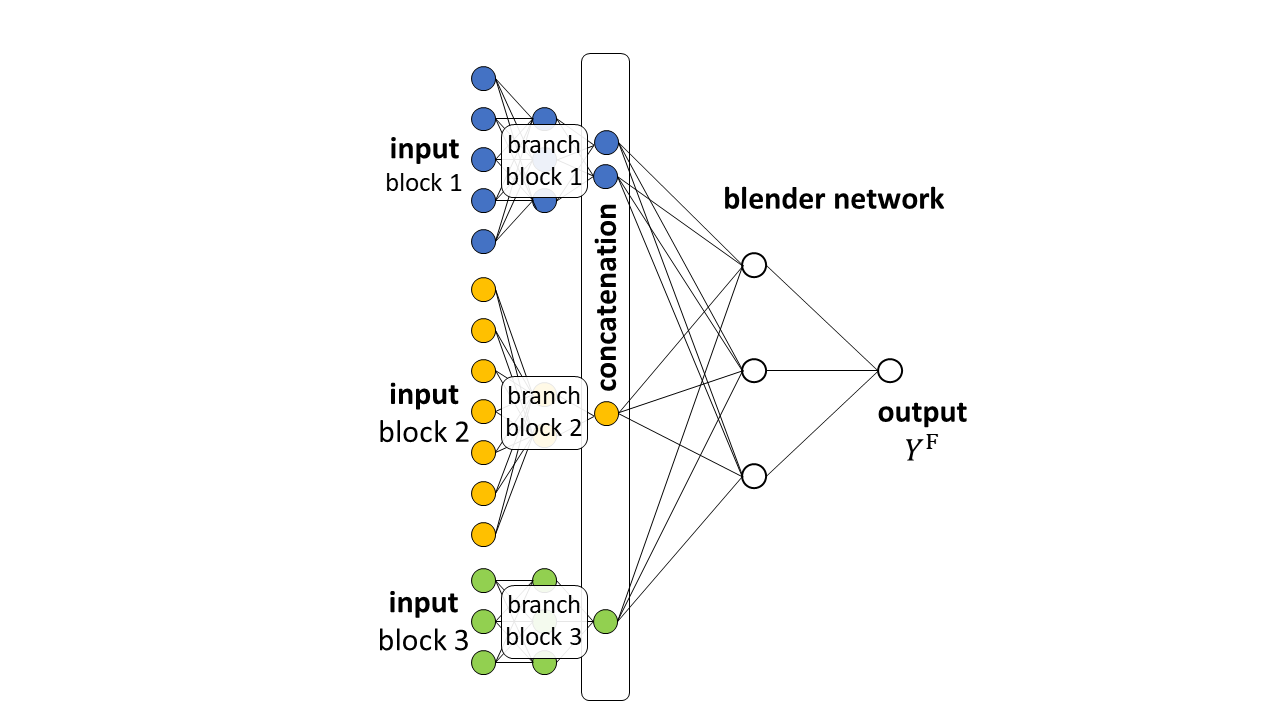}
    \caption{M-ANN architecture.}
    \label{fig:feed_forward_net}
\end{wrapfigure}

In order to exploit the internal structure of the block-wise data in neural networks, a multiblock ANN (M-ANN) architecture is used. As depicted in Figure \ref{fig:feed_forward_net}, the M-ANN consists of a separate input branch for each block, a concatenation layer to merge information from all branches, and a blender network to map the information to the model output. The architecture allows for any type of network layer, depth, activation, or other network parameters, including the special case where the concatenation layer equals the input layer (block branches of depth 0).

Ranking individual features by their importances (feature importance ranking, FIR) has been studied for different types of ANNs \cite{ghorbani2019interpretation, yu2018nisp, wojtas2020feature}. An extensive evaluation \cite{hooker2019benchmark} showed that versions of the variational gradient method (VarGrad) \cite{vargrad,vargrad2} outperformed competitors such as guided backprop and integrated gradients.
For BIR, however, a combination of features in one block may accumulate a larger amount of information than each feature separately due to informative non-linear relations between features.
Hence, using FIR might oversimplify the problem of measuring block importance since interactions between features of the same block are disregarded.
Nevertheless, the strategy of reducing BIR to a simple summary metric (sum, mean, max) over FIR scores is considered in our evaluation.

A related problem to FIR is feature selection, where the input dimensionality is reduced to the most influential features as part of the preprocessing. Feature selection is widely studied in ANNs. Furthermore, specialized feature selectors can account for block structures like UBayFS~\cite{ubayfs} or groupLasso~\cite{grouplasso}. Conceptually, these feature selectors aim to improve model performance and classify an entire block as important/unimportant in a binary way before model training. In contrast, our BIR problem is considered a post-processing procedure, focusing on analyzing the model after training without influence on the model performance.\footnote{BIR may be used for block feature selection if deployed as filter method---however, this aspect is beyond the scope of the present work.}

This study presents and discusses three distinct approaches to quantify the importance of feature-blocks in M-ANNs.
While exploiting the flexibility of ANNs and their capacities to learn complex underlying patterns in the data, the discussed methods aim to deliver insights into the trained network's dependence structure on the distinct input blocks and thereby foster model explainability.
We propose three paradigms for BIR: a block is considered as important if
\begin{enumerate}
    \item it consists of features with high FIR scores (composition strategy), or
    \item it explains a large part of the network output (knock-in strategy), or
    \item its removal significantly changes the network output (knock-out strategy).
\end{enumerate}

We evaluate and discuss the proposed paradigms in a simulation study and present two case studies on real-world datasets, where the behaviors of the proposed ranking strategies become apparent.

In the following, bold letters denote vectors and matrices; non-bold letters denote scalars, functions or sets. Square brackets denote index sets $[n] = \{1,\dots,n\}$.

\section{Block Importance Ranking Methods}

We assume data input $\bm{x}$ from some feature space $D\subset \mathbb{R}^N$, $N\in \mathbb{N}$, following a probability distribution $\bm{X}\sim P_{\bm{X}}$, and a univariate target variable $y\in T\subset \mathbb{R}$ following a probability distribution $Y\sim P_{Y}$. Given training data $(\bm{x},y)\in D_{\text{train}}\times T_{\text{train}} \subset D\times T$, model parameters $\bm{w}\in W \subset \mathbb{R}^M$, $M\in \mathbb{N}$, are trained with respect to some loss term $e:D\times T\rightarrow\mathbb{R}^{+}$, $$\bm{w}^{\star} = \underset{\bm{w}\in W}{\min} ~e\left(f_{\bm{w}}(\bm{x}),y\right),$$
where the ANN is a function $f_{\bm{w}}:D\rightarrow T$ given weights $\bm{w}$.

In an M-ANN architecture, see Fig. \ref{fig:feed_forward_net}, the block structure of the model input is represented by a direct sum of subspaces $D = \bigoplus\limits_{b=1}^{B}D_b$, each corresponding to one block $b\in[B]$ with dimension $N_b = \text{dim}(D_b)$, $N=\sum\limits_{b=1}^{B}N_b$. Each block enters a distinct branch of the network that processes the block input. Afterwards, the outputs of all branches are merged in a concatenation layer, which consists of $n_b$ nodes associated with each block $b$, respectively. A so-called blender network $f_{\bm{w}}^{\text{blender}}$ connects the concatenation layer to the network output. Network training is performed using backpropagation, where all block branches and the blender network are trained simultaneously in an end-to-end manner.

\subsection{Composite strategy}
Our first paradigm composes block importance measures from FIR in a direct way. As a prototype of state-of-the-art FIR methods, we use VarGrad\cite{vargrad}. VarGrad builds on the idea that variations of an important feature provoke measurable variations in the output. Under the assumption that features are on a common scale, we estimate the gradient of the function $f_{\bm{w}}$ with respect to each feature by adding small random perturbations in the input layer. A large variance in the gradient indicates that the network output depends strongly on a feature, i.e., the feature is important. We denote the importance of feature $n\in[N]$ as quantified by VarGrad, by $\alpha_n\in \mathbb{R}^+$.

To translate the feature-wise importance measure to feature-blocks in M-ANNs, we deploy a summary metric $\varphi$ over all single-feature importances in a block $b\in [B]$. Thus, block importances $\gamma_{\varphi}^{(b)}$ are defined as
\begin{equation}
    \gamma_{\varphi}^{(b)} = \varphi(\alpha_1^{(b)},\dots,\alpha_{N_b}^{(b)}),
\end{equation}
where $\alpha_n^{(b)}$ denotes the $n^{\text{th}}$ feature associated with the $b^{\text{th}}$ block. Intuitive choices for $\varphi$ are either the sum, mean, or maximum operator, denoted as $\varphi_{\text{sum}}$, $\varphi_{\text{mean}}$, or $\varphi_{\text{max}}$, respectively. Rankings based on mean and sum are equal, if all blocks contain the same number of features. Operators $\varphi_{\text{sum}}$ and $\varphi_{\text{mean}}$ accumulate the individual feature importances: a block with multiple features of high average importances is preferred over blocks with few top features and numerous unimportant features. In contrast, $\varphi_{\text{max}}$ compares the top-performing features out of each block, while neglecting all other's contributions.
Statistical properties of block importance quantifiers implementing the composite strategy are transmitted from (i) the feature importance ranking method and (ii) the summary metric. Since this approach cannot capture between-feature relations, potentially impacting the importance of a block, two other paradigms are suggested.

\subsection{Knock-in strategy}

The knock-in strategy is inspired by work on the information bottleneck~\cite{geiger19}, demonstrating that node activations can be exploited for model interpretation in ANNs. In the concatenation layer of the M-ANN (Fig. \ref{fig:feed_forward_net}), where information from the blocks enters the blender network, activations are of particular importance since they represent an encoding of the block information. 
When passing model input $\bm{x}$ through the network, we denote the activation of the $n{\text{th}}$ node associated with block $b\in[B]$ in the concatenation layer by $c_{b,n}(\bm{x})$, $n\in [n_b]$. 
The average activation of the $n$th node in block $b\in[B]$ across all training data $\bm{x}\in D_{\text{train}}$ is denoted by $\bar{c}_{b,n}$.

For BIR, we compute a pseudo-output by passing data of only one block $b$ through the network. For this purpose, we introduce a pseudo-input $\bm{v}^{(b)}(\bm{x})$ as
\begin{equation}
v_{b',n}^{(b)}(\bm{x}) = \left\{\begin{array}{ll} c_{b',n}(\bm{x}) & \text{if}~b' = b \\ \bar{c}_{b',n} & \text{otherwise,}\end{array}\right.
\label{eq:KI}
\end{equation}
where $b'\in[B]$, and $n\in[n_b]$. By propagating pseudo-input $\bm{v}^{(b)}(\bm{x})$ through the blender network, we obtain the pseudo-output $f_{\bm{w}}^{\text{blender}}(\bm{v}^{(b)}(\bm{x}))$.
The main assumption behind the knock-in strategy is that high agreement between output $f_{\bm{w}}(\bm{x})$ and pseudo-output $f_{\bm{w}}^{\text{blender}}(\bm{v}^{(b)}(\bm{x}))$ indicates a high importance of block $b$, since information from $b$ is sufficient to recover most of the model output. In contrast, a large discrepancy between the two quantities indicates low explanatory power of the block $b$, and thus, a lower block importance. The concept to generate knock-in pseudo-outputs is illustrated in Figure \ref{fig:knock_in}.

We implement the knock-in concept via the mutual information (MI)\cite{cover2012elements}, an information-theoretic measure to quantify the level of joint information between two discrete random variables $Z$ and $Z'$, defined as
$$ \text{MI}(Z,Z') = \sum\limits_{z}\sum\limits_{z'} p_{Z,Z'}(z,z') \log_2\left(\frac{p_{Z,Z'}(z,z')}{p_Z(z)p_{Z'}(z')}\right).$$
If $Z$ and $Z'$ are independent, $\text{MI}(Z,Z')$ is $0$. Otherwise, $\text{MI}(Z,Z')$ is positive, where a high value indicates a large overlap in information. To quantify the joint and marginal distributions of continuous variables $Z$ and $Z'$, two-dimensional and one-dimensional histograms can be used as non-parametric estimators for $p_{Z,Z'}$, $p_{Z}$, and $p_{Z'}$, respectively. We denote the number of equidistant histogram bins along each axis by $\ell\in\mathbb{N}$. It follows from the properties of entropy\cite{cover2012elements} that an upper bound to $\text{MI}(Z,Z')$ is given by $\log_2(\ell)$.

As shown in Figure \ref{fig:knock_in}, the random variable of (full) model output, $Y^{\text{F}} = f_{\bm{w}}(\bm{X})$, and the random variable of the pseudo-output with respect to block $b$, $Y^{(b)} = f_{\bm{w}}^{\text{blender}}(\bm{v}^{(b)}(\bm{X}))$, where $\bm{X}$ follows the input distribution $P_{\bm{X}}$, are used to measure knock-in (KI) block importance as
\begin{equation}
    \gamma_{\text{KI}}^{(b)} = \frac{\text{MI}(Y^{\text{F}}, Y^{(b)})}{\log_2(\ell)}.
\end{equation}

\begin{minipage}{0.45\textwidth}
    \centering
    \includegraphics[width = 0.4\textwidth, trim=50 0 25 0, clip]{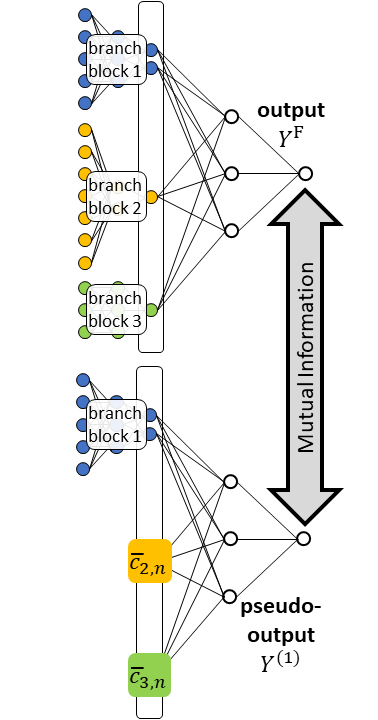}
    \captionof{figure}{Knock-in strategy: Pseudo-outputs for feature-block $b=1$ are generated by activating block $b$, while imputing averaged activations for all other blocks.\label{fig:knock_in}}
\end{minipage} \qquad
\begin{minipage}{0.45\textwidth}
    \centering
    \includegraphics[width = 0.4\textwidth, trim=50 0 25 0, clip]{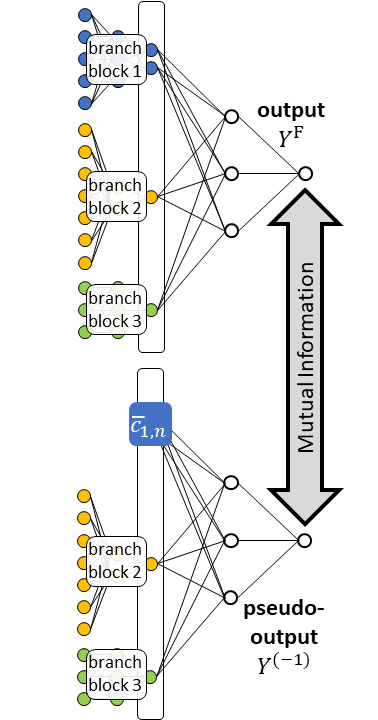}
    \captionof{figure}{Knock-out strategy: Pseudo-outputs are generated by activating all but one blocks $b=1$, while imputing averaged activations for all other blocks.\label{fig:knock_out}}
\end{minipage}

\subsection{Knock-out strategy}
The knock-out paradigm is an ablation procedure where one block at a time is removed from the model in order to measure the impact of the remaining blocks. We pursue a similar approach as in the knock-in paradigm and specify knock-out pseudo-inputs $\bm{v}^{(-b)}(\bm{x})$ as
\begin{equation}
v_{b',n}^{(-b)}(\bm{x}) = \left\{\begin{array}{ll} \bar{c}_{b',n} & \text{if}~b' = b \\ c_{b',n}(\bm{x}) & \text{otherwise,}\end{array}\right.
\label{eq:KO}
\end{equation}
for an arbitrary block $b\in[B]$. Thus, the definition in Eq. \ref{eq:KO} represents an opposite behavior of Eq. \ref{eq:KI} in the knock-in case. In analogy to the knock-in notation, we denote the random variable of pseudo-outputs with respect to $\bm{v}^{(-b)}$ as $Y^{(-b)} = f_{\bm{w}}^{\text{blender}}(\bm{v}^{(-b)}(\bm{X}))$. 
The knock-out concept is illustrated in Figure \ref{fig:knock_out}.
In contrast to knock-in, we assume that leaving out block $b$ having relevant impact on the final output delivers a more dissimilar pseudo-output to the full output since relevant information is lost. Removing an unimportant block preserves the relevant information and delivers a pseudo-output similar to the full output.

Finally, we define the importance of block $b\in[B]$ with respect to the knock-out strategy (KO) as
\begin{equation}
    \gamma_{\text{KO}}^{(b)} = \frac{\log_2(\ell) - \text{MI}(Y^{\text{F}}, Y^{(-b)})}{\log_2(\ell)}.
\end{equation}
For both, KI and KO, importance scores $\gamma_{\text{KI}}^{(b)}$ and $\gamma_{\text{KO}}^{(b)}$ are bounded between $0$ (unimportant block) and $1$ (important block).

\section{Experiments}
As a proof of concept, we conduct two experiments to assess BIR in M-ANNs. The first experiment involves four simulated, non-linear regression problems, where our simulation setup delivers information on the ground truth block importances. This experiment verifies that our suggested measures can identify the ground truth block rankings, defined by their corresponding paradigms. Real-world datasets are evaluated in two case studies in experiment 2, where no exact ground truth block ranking is available. Instead, we compare BIR strategies to each other.

\subsection{Simulation experiment}
We simulate a synthetic datasets along with 6 distinct target functions, denoted as setups S1a-S1c and S2a-S2c. The dataset consists of $N=256$ features, divided randomly into $B=8$ blocks (B1-B8) à $N_b=32$ features. The sample size is set to $\vert D_{\text{train}}\vert = 10000$ and $\vert D_{\text{test}}\vert = 10000$. All features are simulated from a multivariate normal distribution with mean vector $\bm{\mu}=\bm{0}$ and a randomized covariance matrix $\bm{\Sigma}$; hence a non-trivial correlation structure is imposed.
\footnote{Code and details on simulation and network architecture are available at \url{https://github.com/annajenul/Block_Importance_Quantification}}

Setups S1a-S1c and S2a-S2c differ in the parameters used to compute the non-linear target variable $y$, which is simulated via a noisy linear combination of the squared features with coefficient matrix $\bm{\beta}^{(b)}\in\mathbb{R}^{N_b\times N_b}$, given as

\begin{minipage}{0.4\textwidth}
  \begin{align*}
    &y = \underbrace{\sum\limits_{b=1}^8 \bm{x}^T \bm{\beta}^{(b)}\bm{x}}_{g(\bm{x})}+ \varepsilon_{\text{noise}},~\text{where} \\ &\varepsilon_{\text{noise}}\underset{\text{i.i.d.}}{\sim}\mathcal{N}\left(0,\sigma_{\text{noise}}^2\right),~\text{and}
  \end{align*}
\end{minipage}
\begin{minipage}{0.6\textwidth}
  \begin{equation}
        \bm{\beta}^{(b)} = 
            \left(
             \begin{array}{cccc|ccc}
                \beta_{\text{imp}} & 0 & \multirow{2}{*}{\vdots} & 0 & 0 & \multirow{2}{*}{\vdots} & 0\\
                \beta_{\text{int}} & \beta_{\text{imp}}  & & 0 & 0 & & 0\\
                \multicolumn{2}{c}{\cdots} & \ddots & &  \\
                \beta_{\text{int}} & \beta_{\text{int}} &  & \beta_{\text{imp}} & 0 &  & 0 \\
                \hline
                0 & 0 &  &0 & 0 & &0\\
                \multicolumn{2}{c}{\cdots} & & & & \ddots &  \\
                0 & 0 & & 0 & 0 &  & 0
             \end{array}
            \right).
  \end{equation}
\end{minipage}

The matrix $\bm{\beta}^{(b)}\in\mathbb{R}^{N_b\times N_b}$ contains an $N_{\text{imp}}\times N_{\text{imp}}$ quadratic sub-matrix consisting of coefficients $\beta_{\text{imp}}$ of important features, i.e. features with relevant contribution to the target, and interactions $\beta_{\text{int}}$. The noise parameter $\sigma_{\text{noise}}$ is set to 10\% of the standard deviation of the linear combination $g(\bm{x})$ across the generated samples $\bm{x}$.\footnote{Due to the randomized correlation matrix of the feature generation, unimportant features may be correlated with important features, as well as with the target $y$.}
As shown in Tab. \ref{tab:beta_values}, block importances are varied between the setups and  as follows
\begin{itemize}
    \item S1a: varying coefficients of important features, but constant counts;
    \item S1b: varying counts of important features, but constant coefficients; 
    \item S1c: varying counts and coefficients of important features;
    \item S2a-S2c: same as S1a-S1c, but with interaction terms between features.
\end{itemize}

\begin{table}[t]
    \centering
    \caption{Specifications for matrix $\bm{\beta}^{(b)}$: block importance is steered via count $N_{\text{imp}}$, coefficient $\beta_{\text{imp}}$, and interaction $\beta_{\text{int}}$ of the important features.
    }\label{tab:beta_values}
    \begin{tabular}{l | ccc | ccc | c | ccc |ccc}
        \toprule
         \multirow{2}{*}{\diagbox[innerrightsep = 0.2cm]{setup}{block}}
         & \multicolumn{3}{c|}{B1} & \multicolumn{3}{c|}{B2} & \dots & \multicolumn{3}{c|}{B7} & \multicolumn{3}{c}{B8} \\ 
         & $N_{\text{imp}}$ & $\beta_{\text{imp}}$ & $\beta_{\text{int}}$ & $N_{\text{imp}}$ & $\beta_{\text{imp}}$ & $\beta_{\text{int}}$ & \dots & $N_{\text{imp}}$ & $\beta_{\text{imp}}$ & $\beta_{\text{int}}$ & $N_{\text{imp}}$ & $\beta_{\text{imp}}$ & $\beta_{\text{int}}$\\
        \midrule
        S1a & 2 & 7 & 0 & 2 & 6 & 0 &\dots & 2 & 1 & 0 & 0 & 0 & 0\\
        S1b & 7 & 2 & 0 & 6 & 2 & 0 &\dots & 1 & 2 & 0 &  0 & 0 & 0\\
        S1c & 1 & 7 & 0 & 2 & 6 & 0 &\dots & 7 & 1 & 0 & 0 & 0 & 0\\
        \midrule
        S2a & 2 & 7 & 1 & 2 & 6 & 1 &\dots & 2 & 1 & 1 & 0 & 0 & 1\\
        S2b & 7 & 2 & 1 & 6 & 2 & 1 &\dots & 1 & 2 & 1 &  0 & 0 & 1\\
        S2c & 1 & 7 & 1 & 2 & 6 & 1 &\dots & 7 & 1 & 1 & 0 & 0 & 1\\
        \bottomrule
    \end{tabular}
\end{table}

\begin{figure}
    \centering
    \includegraphics[width = 0.6\textwidth]{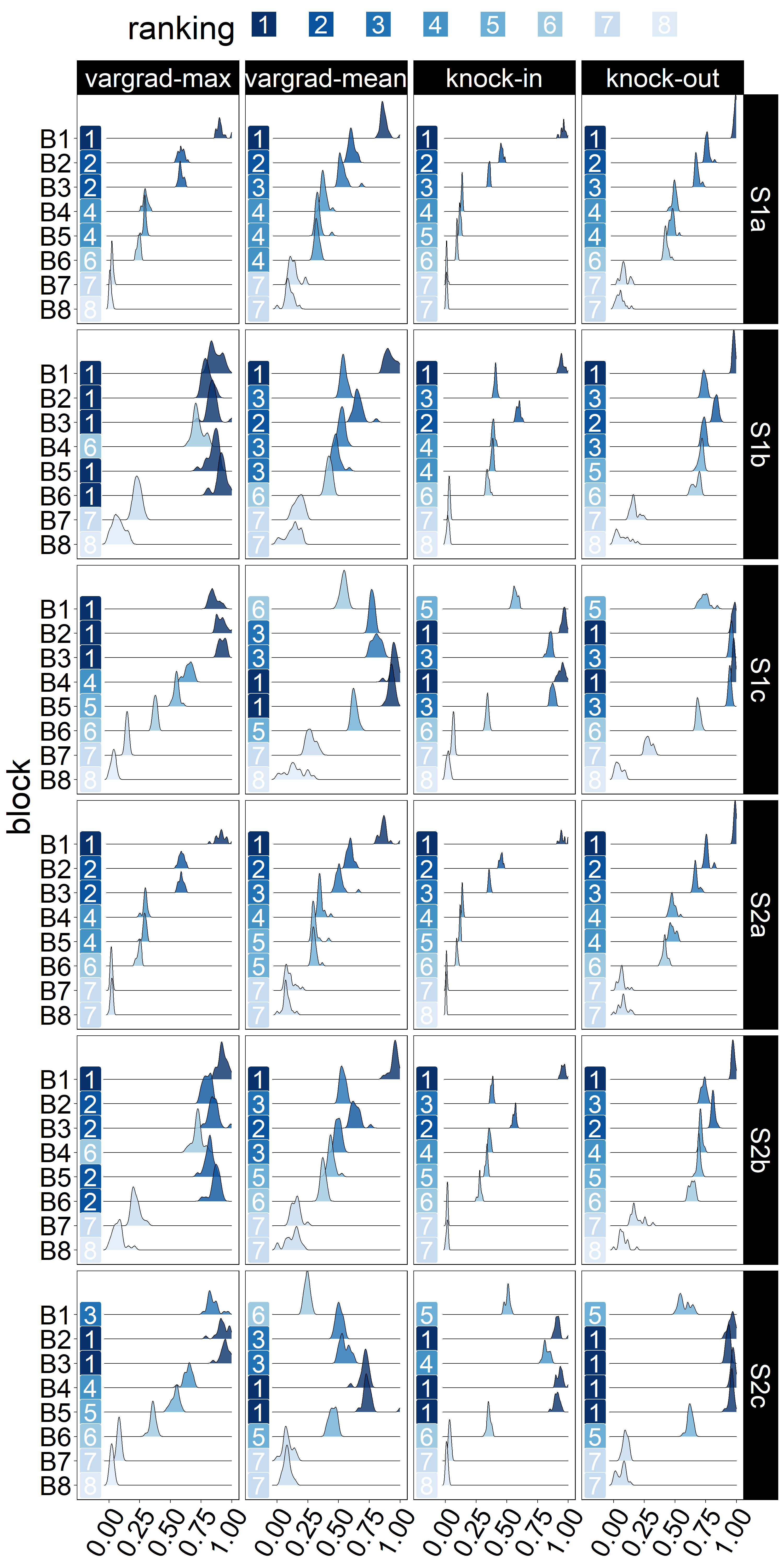}
    \caption{Distributions of the normalized BIR scores across model runs. Rankings are indicated by colors and refer to significant group differences based on a pairwise Wilcoxon-test (significance level $0.01$).}
    \label{fig:sim_plots}
\end{figure}

For each setup, we trained the described M-ANN model in 30 independent runs with distinct weight initializations after data standardization. Since BIR methods are deployed post-hoc and assume a model with appropriate performance, runs with poor performances (R2 $<$ 0.8) were excluded from the analysis after outlier removal, the number of model runs in the analysis was 20 (S1a, S1b, S2a, S2b), 18 (S1c), and 19 (S2c), respectively. The remaining models achieved an average performance of $\geq 0.9$ (R2 score) and $\leq 0.2$ (RMSEIQR: root mean squared error scaled by inter-quartile range) on the test set.

For evaluation, importance scores across all model runs were tested for significant differences using a pairwise Wilcoxon-test with Bonferroni correction. If the p-value in a comparison between two blocks was above a significance level of $0.01$, both were counted as tie. Fig. \ref{fig:sim_plots} illustrates the distributions of BIR scores after min-max-normalization by setup and method, along with rankings (colors) based on significant group differences. The intrinsic ranking in dataset S1a was discovered by all methods. In dataset S1b, knock-in, knock-out and VarGrad-mean identified the ranking by underlying important feature counts $N_{\text{imp}}$, while VarGrad-max failed to deliver a significant distinction between blocks with higher counts of important features. For dataset S1c, VarGrad-max mostly ranked by underlying $\beta_{\text{imp}}$ and ignored $N_{\text{imp}}$, while knock-in, knock-out and VarGrad-mean delivered trade-offs between counts $N_{\text{imp}}$ and coefficients $\beta_{\text{imp}}$ of important features. In setups S2a, S2b, and S2c with between-feature interactions, the same rankings as in the according setups S1a, S1b, and S1c could be obtained by all methods with negligible deviations. Hence, we conclude that all metrics remain stable in more complex scenarios.

We further validated the paradigms by comparing the results to their corresponding ground truth block importances, determined by the real coefficients in the simulation setup. For the composite max and mean paradigms, the corresponding maxima and means over $\bm{\beta}^{(b)}$, were used as references. Ground truth importances for knock-in (KI) and knock-out (KO) were based on the explained variances of the single block $b$ in the underlying linear combination, given as
\begin{equation*}
    KI_b = \mathbb{E}\left(y - \left(\bm{x}^{(b)}\right)^T \bm{\beta}^{(b)}\bm{x}^{(b)}\right), ~\text{and}~
    KO_b = \mathbb{E}\left(y - \sum_{\substack{b'=1 \\ b'\neq b}}^8 \left(\bm{x}^{(b')}\right)^T \bm{\beta}^{(b')}\bm{x}^{(b')}\right),
\end{equation*}
where $\bm{x}^{(b)}$ denotes projection of input $\bm{x}$ on the subspace of block $b$, $D_b$. The comparison between the rankings based on (average) predicted importance scores and ground truth rankings was made using Spearman's correlation coefficient, see Tab. \ref{tab:results_sim}. With two exceptions, all correlation values were at a high level, indicating that our methods accurately predicted the ground truth. Spearman's correlation coefficient is not representative in S1a with respect to the maximum metric since the ground truth ranking is equal for blocks B1-B7. In S2c VarGrad-mean is distracted by decreasing $\beta_{\text{imp}}$, although underlying block importances are in increasing order with respect to the mean metric.

\begin{table}[t]
    \centering
    \caption{Averaged Spearman's rank correlation coefficients comparing each ranking to the ground truth BIR for each paradigm across model runs. Standard deviations were $\leq 0.03$ for S1a, S1b, S2a and S2b, and $\leq 0.06$ for S1c and S2c.}
    \label{tab:results_sim}
    \begin{tabular}{l | c | c| c | c | c | c}
    \toprule
    \diagbox[innerrightsep = 0.2cm]{paradigm}{dataset} & S1a & S1b & S1c & S2a & S2b & S2c \\
    \midrule
     composite (VarGrad-mean) & $0.98$ & $0.95$ & $0.96$ &  $0.97$ & $0.95$ & $-0.40$ \\
    composite (VarGrad-max) & $0.97$ & $0.58$& $0.93$ & $0.98$ & $0.58$ &  $0.91$\\
    knock-in & $0.99$ & $0.98$ &$0.85$ & $0.97$ & $0.99$ &  $0.89$ \\
    knock-out & $0.99$ & $0.98$ & $0.81$ & $0.99$ & $0.99$ &  $0.89$ \\
    \bottomrule
    \end{tabular}
\end{table}

\subsection{Real-world experiment}

Since verification on simulated data showed that the presented approaches match the ground truth according to their paradigms, we deployed the methods on two real-world datasets, where underlying block importance is unknown. Prior to analysis both datasets were standardized on the trained data. Again, we trained 30 independent models runs.

\begin{figure}[t]
    \centering
    \begin{subfigure}{0.65\textwidth}
        \includegraphics[width = \textwidth]{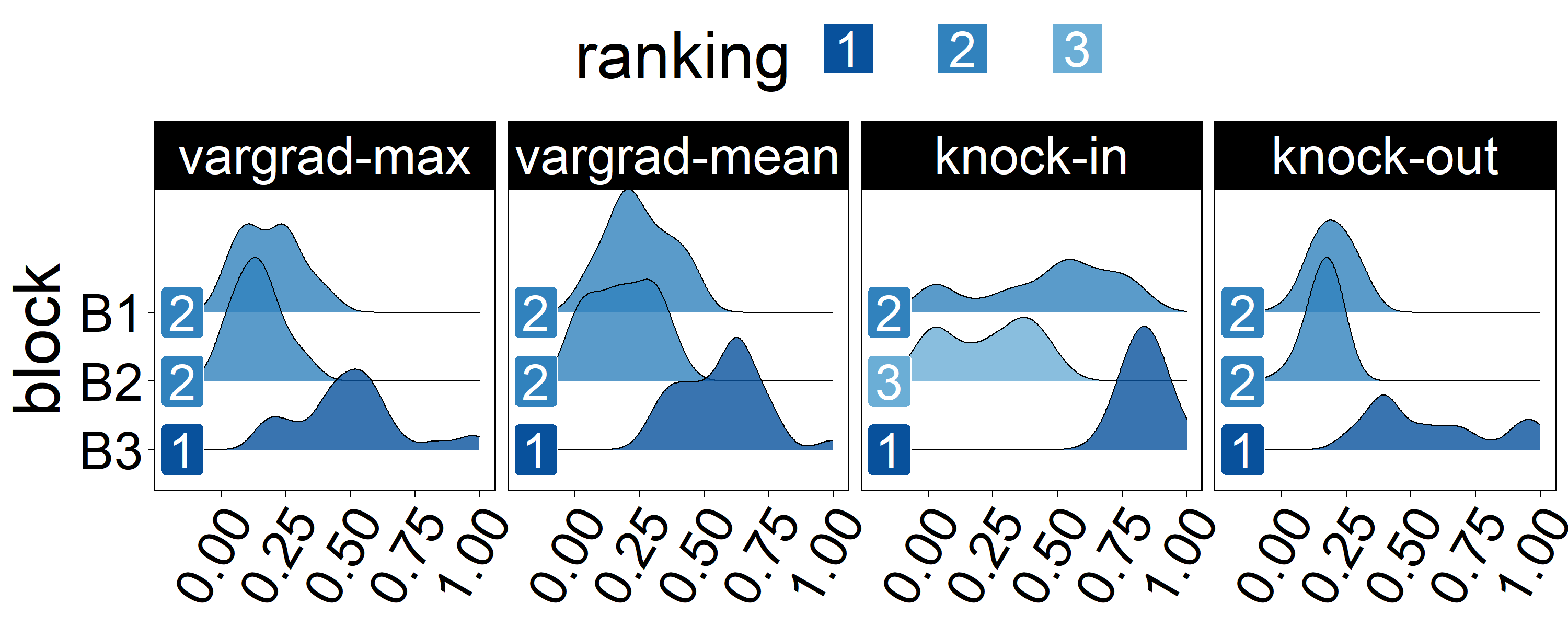}
        \caption{Breast Cancer Wisconsin dataset.}
        \label{fig:real-world-bcw}
    \end{subfigure}
    \begin{subfigure}{0.65\textwidth}
         \includegraphics[width = \textwidth]{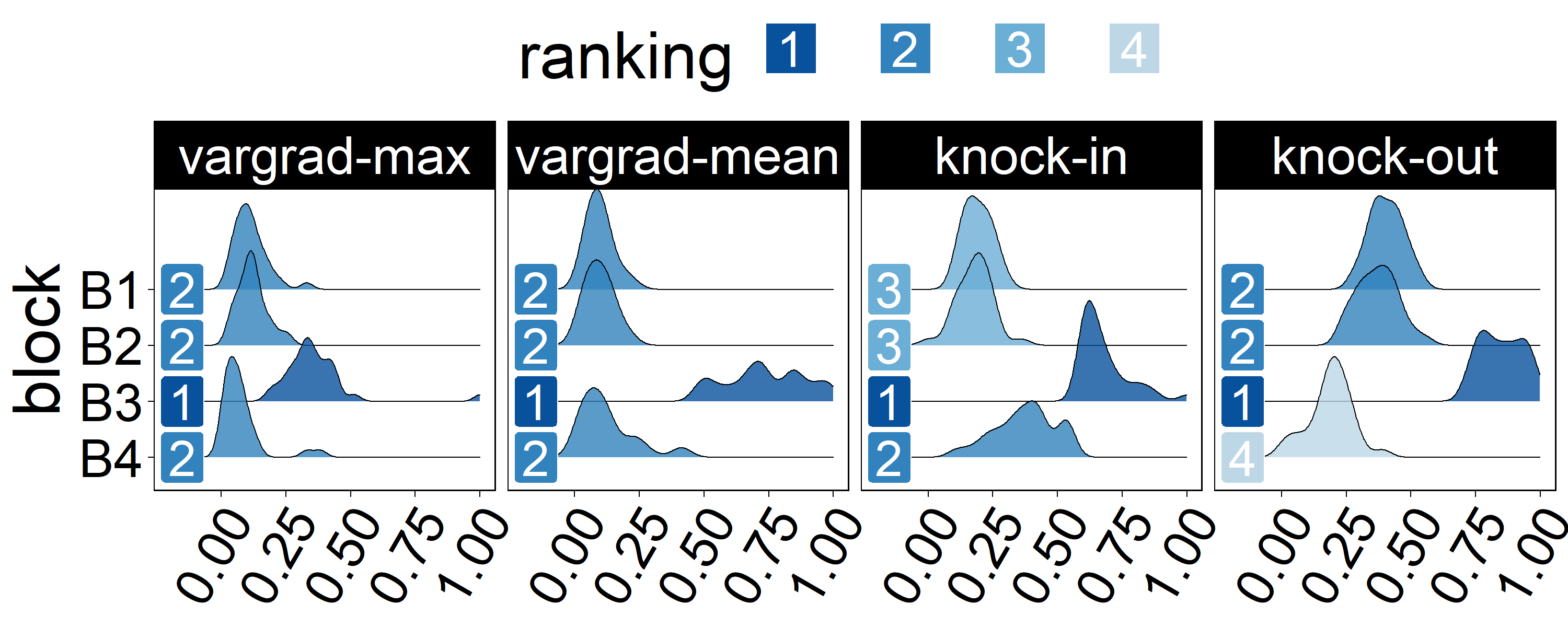}
         \caption{Servo dataset.}
         \label{fig:real-world-servo}
    \end{subfigure}
    \caption{Distributions of normalized BIR scores in experiment 2 across model runs.}
    \label{fig:real-world}
\end{figure}

The Breast Cancer Wisconsin dataset (BCW) \cite{street:bcw} describes a binary classification problem (malignant or benign tumor) and consists of 569 samples (398 train, 171 test) and three blocks with ten features each, representing groups of distinct feature characteristics (mean values, standard deviations, and extreme values of measured parameters). The average performance was 0.95 (accuracy) and 0.96 (F1 score) without outliers. The average scores and rankings delivered by BIR methods are shown in Fig. \ref{fig:real-world-bcw}. All four paradigms discovered that block 3 is dominant, which agrees with previous research on the dataset~\cite{jenul2021:related}. However, knock-in was the only method, which distinguished between the importances of B1 and B2. According to \cite{jenul2021:related}, block B1 contains overlapping information with B3, while B2 is rather non-informative. Thus, the experiment underlines a difference between knock-in and knock-out rankings in the presence of redundancies.

Servo \cite{servo} is a dataset containing 167 samples (120 train, 47 test), a univariate, numeric target variable, and four features, two of which are categorical variables with four levels each, and two are numerical variables. Each feature was assigned its own block. For the two blocks containing categorical features, one-hot encoding was performed, leading to two blocks (B1 and B2) of four binary features, each. Blocks corresponding to numerical features (B3 and B4) contain one feature, each. In the 30 M-ANN model runs, an average performance of 0.21 (RMSEIQR) and 0.87 (R2) was obtained without outliers. Fig. \ref{fig:real-world-servo} shows that for all four paradigms, block B3 was most important. While VarGrad methods delivered a binary ranking, knock-in and knock-out suggested a ranking with 3 and 4 distinct importance levels, respectively---thus, the level of detail was higher in the MI-based rankings compared to VarGrad methods.

\section{Discussion}
Our experiments demonstrated several differences between the proposed strategies.
While the composite strategy evaluates features individually and depends on two user-selected parameters (the feature-wise ranking scheme and the summary metric), the knock-in and knock-out strategies consider each block a closed unit. They require no selection of a summary statistic. MI-based rankings deliver a score in $[0,1]$, while VarGrad has no upper bound. However, the discretization associated with the mutual information calculation may influence the importance scores and, thus, the rankings by knock-in and knock-out. All strategies are applicable for multivariate target variables, as well. However, an $\text{MI}$-based comparison between outputs and pseudo-outputs is prone to suffer from the curse of dimensionality since higher-dimensional probability distributions are compared to each other. On the contrary, the vanishing gradient problem can influence VarGrad in deep ANN architectures. All approaches delivered accurate experimental results, but only knock-in and knock-out provided a consistent ranking of blocks with minor importance in dominant blocks, such as for the servo dataset.

Even though knock-in and knock-out rely on the same concept of assessing pseudo-outputs related to each block, their properties and interpretations differ. The knock-in strategy determines whether a block can deliver a reasonable target description independently from the remaining blocks. This interpretation of block importance evaluates the performance achieved if we reduce the model to solely one block at a time.
In contrast, knock-out quantifies whether the contribution of a block can be compensated by any other block. If two blocks contain redundant information about the target, knock-in delivers high values for both blocks since each block individually has high explanatory power. In contrast, knock-out penalizes redundant blocks since each of them can be removed without loss of information. This property became evident in the BCW experiment, where B3 was dominant but shared overlapping information with B1: knock-in was the only approach that discovered the higher information content in B1 compared to the uninformative B2.

\section{Conclusion}
We have demonstrated three strategies to rank the importance of feature-blocks as post-processing in ANNs with block-wise input structure. The composite strategy, which is a direct generalization of feature-wise importance rankings, provided promising results in most cases, but selecting the correct summary statistic was crucial. Knock-in and knock-out strategies, implemented using an information-theoretic measure on the model outputs, delivered a trade-off between the extremes of maximum and mean feature importance in the composite case. All methods uncovered the true block importance with high accuracy and delivered new insights into the ANN's behavior. Still, computing multiple proposed metrics is advantageous for making informative block ranking decisions.

\bibliographystyle{abbrv}
\bibliography{references}

\end{document}